\documentclass[conference]{IEEEtran}
\IEEEoverridecommandlockouts
\usepackage{cite}
\usepackage{amsmath,amssymb,amsfonts}
\usepackage{algorithmic}
\usepackage{graphicx}
\usepackage{textcomp}
\usepackage{xcolor}
\usepackage{times}
\usepackage{epsfig}

\usepackage{multirow}
\usepackage{makecell}
\usepackage{subfig}

\IEEEpubid{ \makebox[\columnwidth]{978-1-6654-0645-1/21/\$31.00 ~\copyright2021 IEEE\hfill} \hspace{\columnsep}\makebox[\columnwidth]{} } 

\begin{document}

\title{GABOR FILTER INCORPORATED CNN \\FOR COMPRESSION}

\author{\IEEEauthorblockN{Akihiro Imamura}
\IEEEauthorblockA{\textit{Embedded AI R\&D department} \\
\textit{DENSO CORPORATION}\\
Shinagawa, Tokyo, Japan \\
akihiro.imamura.j5f@jp.denso.com}
\and
\IEEEauthorblockN{Nana Arizumi}
\IEEEauthorblockA{\textit{Embedded AI R\&D department} \\
\textit{DENSO CORPORATION}\\
Shinagawa, Tokyo, Japan \\
nana.arizumi.j8j@jp.denso.com}
}

\maketitle

\begin{abstract}Convolutional neural networks (CNNs) are remarkably successful in many computer vision tasks.  However, the high cost of inference is problematic for embedded and real-time systems, so there are many studies on compressing the networks. On the other hand, recent advances in self-attention models showed that convolution filters are preferable to self-attention in the earlier layers, which indicates that stronger inductive biases are better in the earlier layers.  As shown in convolutional filters, strong biases can train specific filters and construct unnecessarily filters to zero.  This is analogous to classical image processing tasks, where choosing the suitable filters makes a compact dictionary to represent features.  We follow this idea and incorporate Gabor filters in the earlier layers of CNNs for compression.  The parameters of Gabor filters are learned through backpropagation, so the features are restricted to Gabor filters.  We show that the first layer of VGG-16 for CIFAR-10 has 192 kernels/features, but learning Gabor filters requires an average of 29.4 kernels.  Also, using Gabor filters, an average of 83\% and 94\% of kernels in the first and the second layer, respectively, can be removed on the altered ResNet-20, where the first five layers are exchanged with two layers of larger kernels for CIFAR-10.
\end{abstract}


\section{Introduction}

In recent years, convolutional neural networks (CNNs) have achieved remarkable success in computer vision tasks, e.g. image classification \cite{krizhevsky2012imagenet}, object detection \cite{girshick2014rich}, semantic segmentation \cite{long2015fully}. 
The networks have grown deeper \cite{szegedy2015going, he2016deep}, and wider \cite{zagoruyko2016wide} to achieve higher accuracy in these tasks.  Hence, the networks are increasing to have more parameters with a high cost of inference operations.  It is a major problem for mobile devices and embedded systems applications, which has a limitation in memory, power, and computation speed.  Also, it is an issue for real-time systems like image classification for web services, which also has a time budget to process inferences. 

Hence, there has been a significant amount of works on model compression, and acceleration \cite{cheng2017survey, cheng2018recent}.  Among which, network pruning from well-trained CNN has been widely studied \cite{lecun1990optimal, hassibi1993second, han2015deep}.  There are different granularities of pruning.  Fine-grained level (e.g., weight level) sparsity gives the highest flexibility and has high compression. However, it also gives unstructured sparsity and needs specialized hardware or/and sparse libraries to harvest the best efficiency from the compression \cite{han2015deep, han2015learning, guo2016dynamic}.  On the contrary, structured pruning does not require them.  Hence, there are increasingly more works on channel-level pruning \cite{he2017channel, liu2017learning, luo2017thinet} and kernel-level pruning \cite{anwar2017structured}.  Note that the channel-level pruning in one layer also makes kernel-level pruning in the next layer. Layer-level pruning is also possible for deep CNNs \cite{huang2016deep, wen2016learning}.  In this work, we also focus on pruning without introducing unstructured sparsity.

The reason why we can prune the weights is that after training, the filters become small enough to be ignored; however, we do not know which to be pruned without training the network \cite{frankle2018lottery}.  Hence, we can assume that unnecessary filters become zero by training with weight decay.  However, the filters might still have redundancy. The success of compressing the network using low-rank decomposition of the convolutional layers \cite{denil2013predicting, zhang2015accelerating} indicates that features required to represent the layer are relatively small. 

The recent advances in self-attention models showed that convolution filters are preferable to self-attention filters in the earlier layers \cite{d2021convit, ramachandran2019stand, srinivas2021bottleneck}, which indicates that stronger inductive biases are better in the earlier layers.  So, additional inductive biases might help choose the right filters in the earlier layers.  Since it has been known that the networks are trained to become the Gabor filters in the earlier layers of CNNs \cite{krizhevsky2012imagenet}, we can insert additional biases by enforcing filters to be the Gabor filters. 

We observed that the first layer of VGG-16 on CIFAR-10 requires an average of 13.2 $7 \times 7 \times 3$ channels of Gabor kernels without accuracy degradation.  On the contrary, the original VGG-16 has 64 $3 \times 3 \times 3$ channels in the first layer. Hence, in the first two layers, it makes 24\% computation and memory reduction.  Even more, an average of 162.6 $7 \times 7$ kernels out of 192  can be removed without accuracy degradation using kernel-level pruning.  Similar behavior is observed for ResNet-20 on CIFAR-10. We extended this idea to the second layer and experimented with altered ResNet-20, where the first five layers are exchanged with two layers of $7 \times 7$ and $5 \times 5$ without shortcut connection.  It shows that 83\% and 94\% of kernels can be removed in the first and second layers, respectively, without accuracy degradation.

Using well-trained CNNs, we can fit and retrain the features by restricting to Gabor filters in the earlier layers.  This way, we can insert stronger inductive biases in the earlier layers and help to choose more specific features.  As with classical feature detection, clear feature prediction can significantly reduce the size of CNNs.  Finding better features and reduce redundant channels is an excellent way to compress CNNs, because it does not create unstructured sparse pruning.  
Furthermore, any other compressing methods can be applied after our channel-level pruning since it does not change the structure of CNNs.  

\section{Related Work}
The Gabor filters are the classical method that can extract local features \cite{jain1997object}, and it is particularly popular because of invariance in scale, rotation, and translation.  Also, the linear convolution of the Gabor filters is a good representation of the simple cells in the human visual cortex \cite{daugman1985uncertainty, jones1987evaluation}, which is a good indication that the early stage of feature extraction of the human visual system also uses Gabor filters. Analogously, the earlier layers of CNNs are similar to Gabor filters \cite{coates2011analysis, krizhevsky2012imagenet}.  Hence, there are several studies on utilizing Gabor filters with CNNs. 

Gabor filters can be used as pre-processing of the input data \cite{yao2016gabor, calderon2003handwritten, kwolek2005face}, or initialization \cite{chang2014robust}.  
\cite{sarwar2017gabor} uses fixed Gabor filters for the first or the second layer to reduce training complexity. \cite{luan2018gabor} incorporated selected Gabor orientation filters to each layer to improve the robustness of the networks.   \cite{yuan2020deep} uses Gabor filters for the interpretability of the networks in applying person re-identification.  They train Gabor filters' parameters but do not learn the scale or the center of the filters since their interests are on the features' interpretability and not to remove the features.   In our work, we also used backpropagation to learn the parameters of Gabor filters in the network to enforce the layer to be the suitable feature detection. 

\section{methodology}
In recent CNNs, kernel sizes are mostly $1 \times 1$ and $3 \times 3$. It is mostly because of computational efficiency.
However, the first layer can have larger filters because the input images are usually one or three channels, and the computational cost in channel size is much smaller.  In fact, there are several popular networks with larger first filters \cite{krizhevsky2012imagenet, he2016deep}.  We could expand the kernels in the first layer for the other CNNs with a smaller kernel. Expanding kernels itself should not change the accuracy drastically. Even though the kernel size becomes larger, if the number of channels compressed, then the total computation cost can be smaller than the original CNNs. Also, a larger kernel size means it covers larger receptive fields, so if the first few layers are only to expand receptive fields, then the layers can also be pruned.  It is more common to use larger kernels in earlier layers for larger images.

Once the first layer's kernel is expanded and trained, we can fit the Gabor filters to the well-trained network.  Then, train Gabor filter parameters using backpropagation.  This way, we can enforce learned features to be Gabor filters.  After retraining, prune channels using the L1 norm criterion as in \cite{li2016pruning}. 


\begin{figure}
  \centering
  \vspace{-1cm}
  \includegraphics[width=8cm]{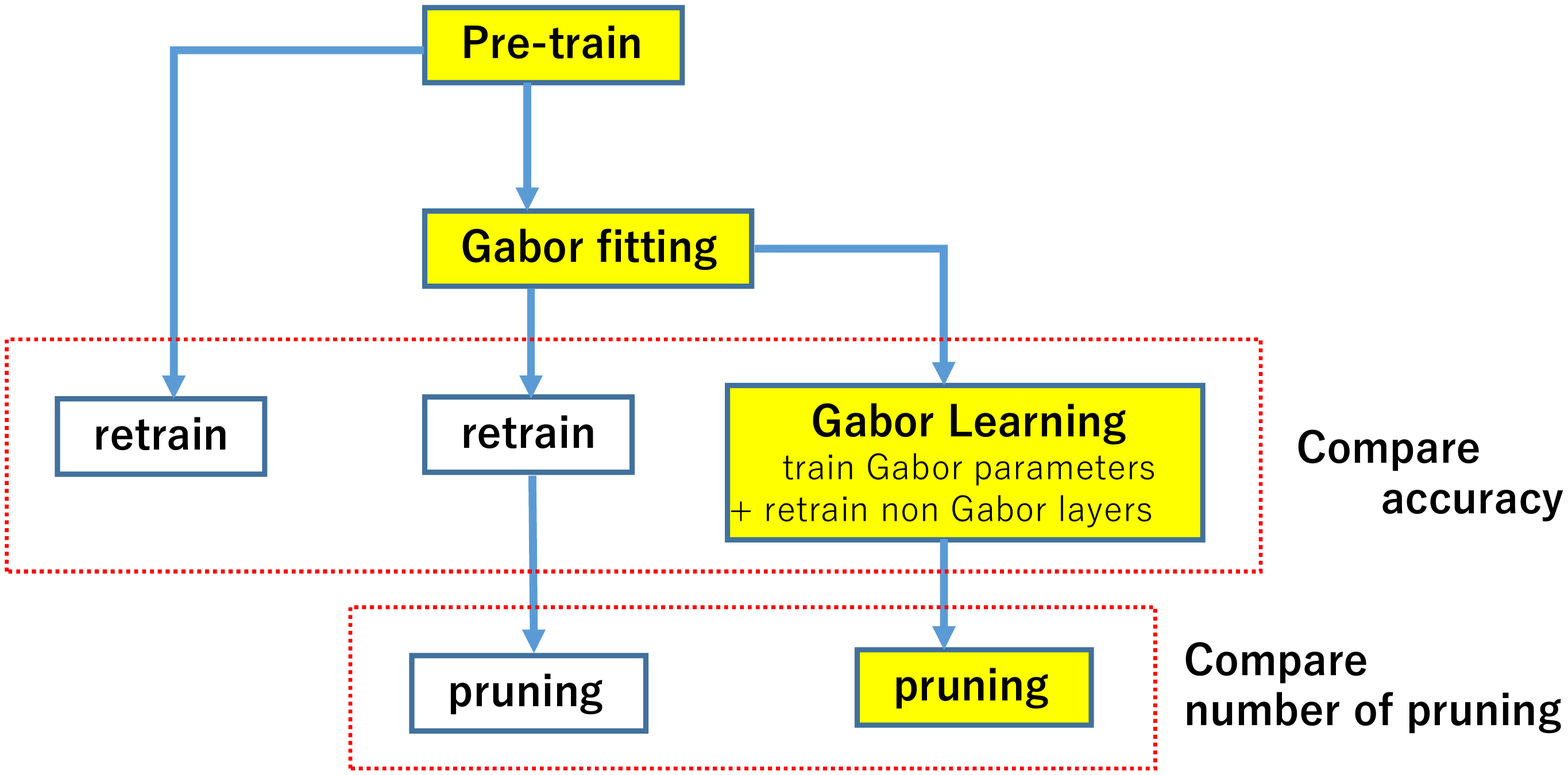}
  \vspace{-1cm}
  \caption{Schematic of pruning methods}
  \label{fig:scheme1}
\end{figure}
Fig. \ref{fig:scheme1} illustrates the schematic of this method in yellow.  We also show the other methods for comparison.  Here, "retrain" does not mean resuming the learning.  We want to compare each method, so we retrain the network with the same hyperparameters, including learning rates.  Even though the network training is converged in the pretraining stage, it is known that the changing learning rate increases accuracy \cite{smith2017cyclical}. 

\subsection{Gabor Fitting}
The Gabor filter \cite{gabor1946theory} is defined by a sinusoidal wave multiplied by the Gaussian function.  
We use the real part 
\begin{equation}
  G = a \, \exp(-\frac{\hat{x}^2 + \gamma \hat{y}^2}{2 \sigma^2})\cos(2\pi \frac{\hat{x}}{\lambda} + \psi),
\end{equation}
where 
\begin{align}
\hat{x} &=& (x-x_0) \cos(\theta) + (y-y_0) \sin(\theta),\\
\hat{y} &=& -(x-x_0) \sin(\theta) + (y-y_0) \cos(\theta),
\end{align}

$\lambda$ is the wavelength of the sinusoidal factor, $\theta$ is the orientation of the normal to the parallel stripes, $\psi$ is the phase offset, $\sigma$ is the standard deviation of the Gaussian envelope, $\gamma$ is the spatial aspect ratio, $a$ is the amplitude, and $(x_0, y_0)$ is the center. Consider the kernels are located at the $(x, y)$ grids from $(1, 1)$ to $(k, k)$, where $k$ is the size of the kernel.

Construct Gabor filters using the following parameters and choose the closest to the original kernel using the L2 norm.

\begin{itemize}
  \item amplitude $a$ to be $-1, -0.5, 0, 0.5, 1$
  \item center $(x_0, y_0)$ to be all the combination between 1 to size of kernel
  \item orientation $\theta$ and phase offset $\psi$ to be $0, \pi/4, \pi/2, 3 \pi/4$
  \item standard deviation of the Gaussian envelope $\sigma$ and wavelength $\lambda$ to be $1, 2, 3, 4, 5$ 
  \item aspect ratio $\gamma$ to be $0.2, 0.4, 0.6, 0.8, 1$    
\end{itemize} 

\subsection{Gabor Learning}
Using backpropagation, we can train the parameters of Gabor filters, $\lambda$, $\theta$, $\psi$, $\sigma$, $\gamma$, $a$, $x_0$, and $y_0$.  
The details of backpropagation are written in \cite{yuan2020deep}.  Contrary to their method, we did not set a range of values of parameters. Also, we add the amplitude and the center of the filters for learnable parameters.  We intend to prune the filters, so amplitude should dissipate, and we can remove filters altogether.  We did not use regularization loss of the Gabor parameter since we did not observe the difference.  We used the Autograd package in PyTorch \cite{paszke2019pytorch} to implement.  For our experiments, the same learning rate is applied to all parameters of Gabor filters.

\section{Experiments}
In CNNs, input feature maps $X_i$ are convolved with channels $C_i$ for $n_{i+1}$ times and produce output feature maps $X_{i+1}$ at the $i$th layer.  The channels $C_i$ has $n_i$ kernels of the size $k_i \times k_i$.  We use shorthand notation of $k_i \times k_i, n_i$ for the channels.  In our experiments, all input images are 3 channels.

We used CIFAR-10 \cite{krizhevsky2009learning} image database, which has $32 \times 32$ natural images. Note, $7 \times 7$ kernels can cover a large part of the image.  There are 5000 training images and 1000 test images.  Standard data augmentation and normalization of input data are also applied.    

We conducted experiments, as shown in Fig. \ref{fig:scheme1} on two networks, VGG \cite{simonyan2014very} and ResNet \cite{he2016deep}.  Both of which are widely popular networks and commonly used for pruning experiments.  We followed VGG-16 from \cite{li2016pruning} and ResNet-20 from \cite{neta_zmora_2018_1297430} for pre-processing, hyper-parameters, and learning schemes.  It has been known that the network training is sensitive to initialization and order of training \cite{krahenbuhl2015data}.  So, we ran all the experiments 5 times and took an average.  We pruned channels and kernels that do not reduce accuracy by more than 0.2\%.  The channels and the kernels are chosen from the minimum according to the L1-norm.  The testing for each case is operated with the same 79 batches with 128 images. 

\subsection{Gabor Fitting of first layer}
For both VGG-16 and ResNet-20, we expanded 3 $\times$ 3 kernels of the first layer to $7 \times 7$.  As for ResNet-20, we remove two layers to fit the receptive field since it only has 0.5\% degradation of accuracy.  We did not do so for VGG-16, because it causes more than 2\% of accuracy loss.  First, we train VGG-16 for 350 Epochs twice with the same hyperparameters and ResNet-20 for 170 Epochs once, then fit the first layer to the Gabor filters.  After retraining using conventional convolution and Gabor learning, we compare the accuracy as shown in Tab. \ref{VGG_accuracy}. Fitting does not degrade the accuracy.

\begin{table}[!tb]
  \renewcommand{\arraystretch}{1.3}
  \caption{Accuracy of Gabor Learning}
\label{VGG_accuracy}
\centering
\begin{tabular}{|c||c|c|}
  \hline
  first layer & $3 \times 3$ & $7 \times 7$\\
  \hline
  \multicolumn{3}{|c|}{VGG-16}\\ 
  \hline  
  pre-train & 93.37 $\pm$ 0.09 & 93.23 $\pm$ 0.13 \\
  \hline
  \hline
  without fit + retrain & 93.44 $\pm$ 0.18 & 93.30 $\pm$ 0.13\\
  \hline
  fit + retrain & - & 93.39 $\pm$ 0.08\\
  \hline
  fit + Gabor learning & - & 93.11 $\pm$ 0.08\\
  \hline
  \multicolumn{3}{|c|}{ResNet-20}\\ 
  \hline  
  pre-train & 91.25 $\pm$ 0.14 & 90.80 $\pm$ 0.07 \\
  \hline
  \hline
  without fit + retrain & 91.89 $\pm$ 0.09 & 91.48 $\pm$ 0.10\\
  \hline
  fit + retrain & - & 91.41 $\pm$ 0.08\\
  \hline
  fit + Gabor learning & - & 91.42 $\pm$ 0.10\\
  \hline
  \end{tabular}
\end{table}

The percentile of prunable kernels and channels are shown on Tab. \ref{Tab:firstResult}. For VGG-16, the percentile of channel pruning is the same as the results with $3 \times 3$ kernels without fitting presented in \cite{li2016pruning}.  Note that for ResNet-20, two layers are also removed, but that is not shown on the table.

\begin{table}[!tb]
  \renewcommand{\arraystretch}{1.3}
  \caption{First layer pruning}
\label{Tab:firstResult}
\centering
\begin{tabular}{|c||c|c|c|c|}
  \hline
  & \multicolumn{2}{c}{VGG-16} & \multicolumn{2}{|c|}{ResNet-20}\\ 
  \hline
  & channel & kernel & channel & kernel \\
  \hline
  retrain & 49\% & 55\% & 14\% & 21\%\\
  \hline
  Gabor learning & 79\% & 91\% & 50\% & 75\% \\
  \hline
\end{tabular}
\end{table}

\begin{figure}
  \centering
  \includegraphics[width=8cm]{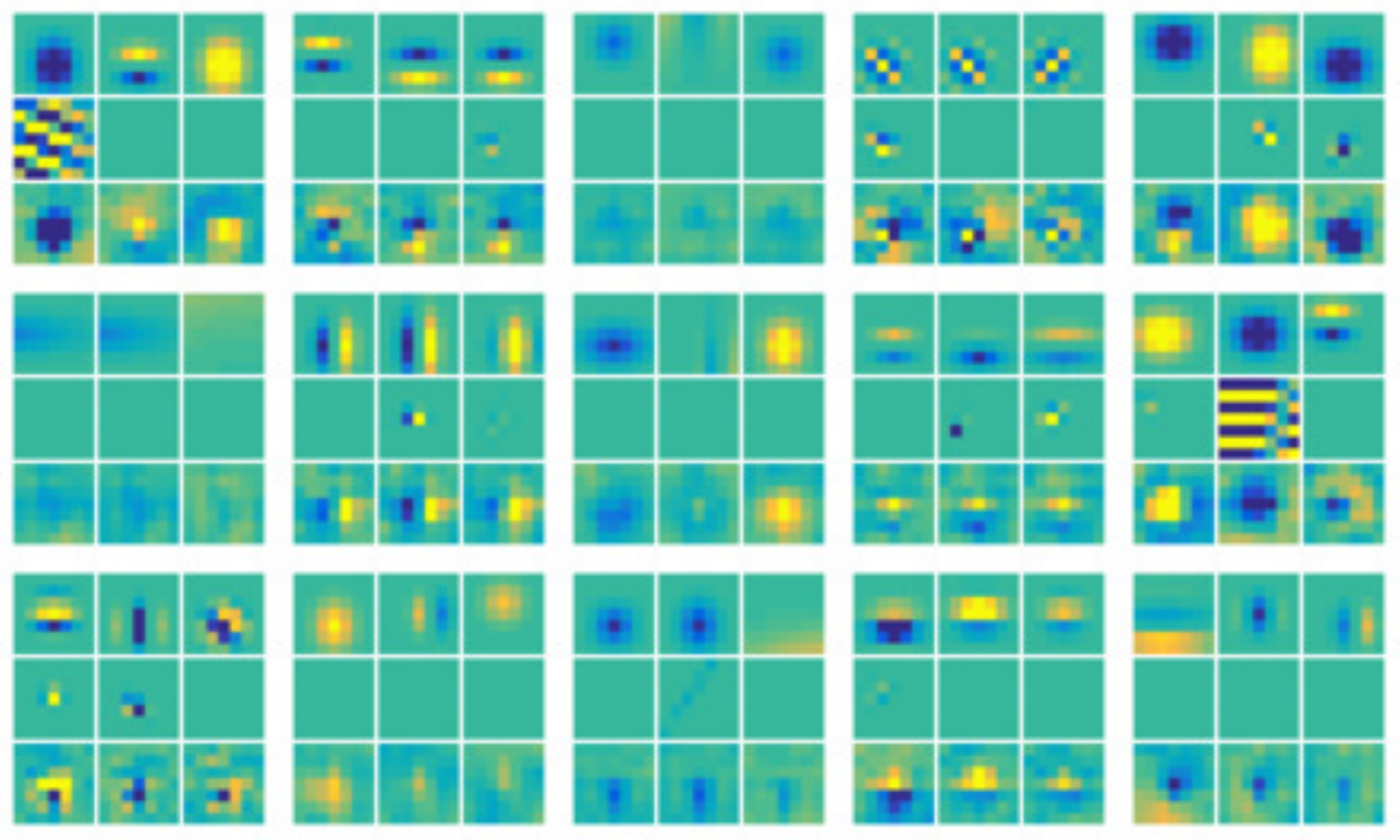}
  \caption{Examples of kernels in the first layer channels of ResNet-20 for both conventional convolutional learning and Gabor learning after Gabor fit.  Inside the 3 $\times$ 3 block, the first row shows after the Gabor fit, the second row shows the Gabor learning after Gabor fit, and the third row shows the conventional convolutional learning after Gabor fit.  Each column shows filters for RGB images.}
  \label{fig:ResNet3x5}
\end{figure}

To show how the kernels were learned, Fig. \ref{fig:ResNet3x5} shows examples of kernels in the first layer channels of ResNet-20 for both conventional convolutional learning and Gabor learning after Gabor fit.  Each kernel is normalized, and the colormap is shown in 6 $\sigma$.  For each block with 3 $\times$ 3 kernels, the first row shows after the Gabor fit, the second row shows the Gabor learning after Gabor fit, and the third row shows the conventional convolutional learning after Gabor fit.  Both of the learnings are starting from the first row.  The columns are corresponding to the 3 input channels of RGB images.  It clearly shows that conventional convolutional learning does not change much from the Gabor fit. Still, many of the kernels after Gabor learning dissipates, and some form different structures.  However, once the kernels are close to zero in initialization, it does not change to other structures. It seems that unnecessary filters dissipate by weight decay. 
 
\subsection{Gabor Fitting of first two layers}

\begin{table}[!t]
  \renewcommand{\arraystretch}{1.3}
  \caption{Accuracy of expanded ResNet-20 after second layer Gabor fitting}
\label{tab:7x7_5x5_2}
\centering
\begin{tabular}{|c|c|c||c|}
  \hline
  &7$\times$7 & 5$\times$5 & accuracy \\
  \hline
  \hline
  \multirow{4}{*}{\makecell{5$\times$5\\fitting}}& retrain & retrain & 91.92 $\pm$ 0.13 \\
  \cline{2-4}
  &Gabor & retrain & 91.88 $\pm$ 0.07\\
  \cline{2-4}
  &retrain & Gabor & 91.70 $\pm$ 0.06\\
  \cline{2-4}
  &Gabor & Gabor & 91.60 $\pm$ 0.42\\
  \hline
  no fitting &Gabor & retrain &  91.96 $\pm$ 0.14 \\
  \hline
  \end{tabular}
\end{table}

\begin{figure}  
  \centering
  \includegraphics[clip, trim=8cm 1cm 2cm 1cm, width=8.5cm]{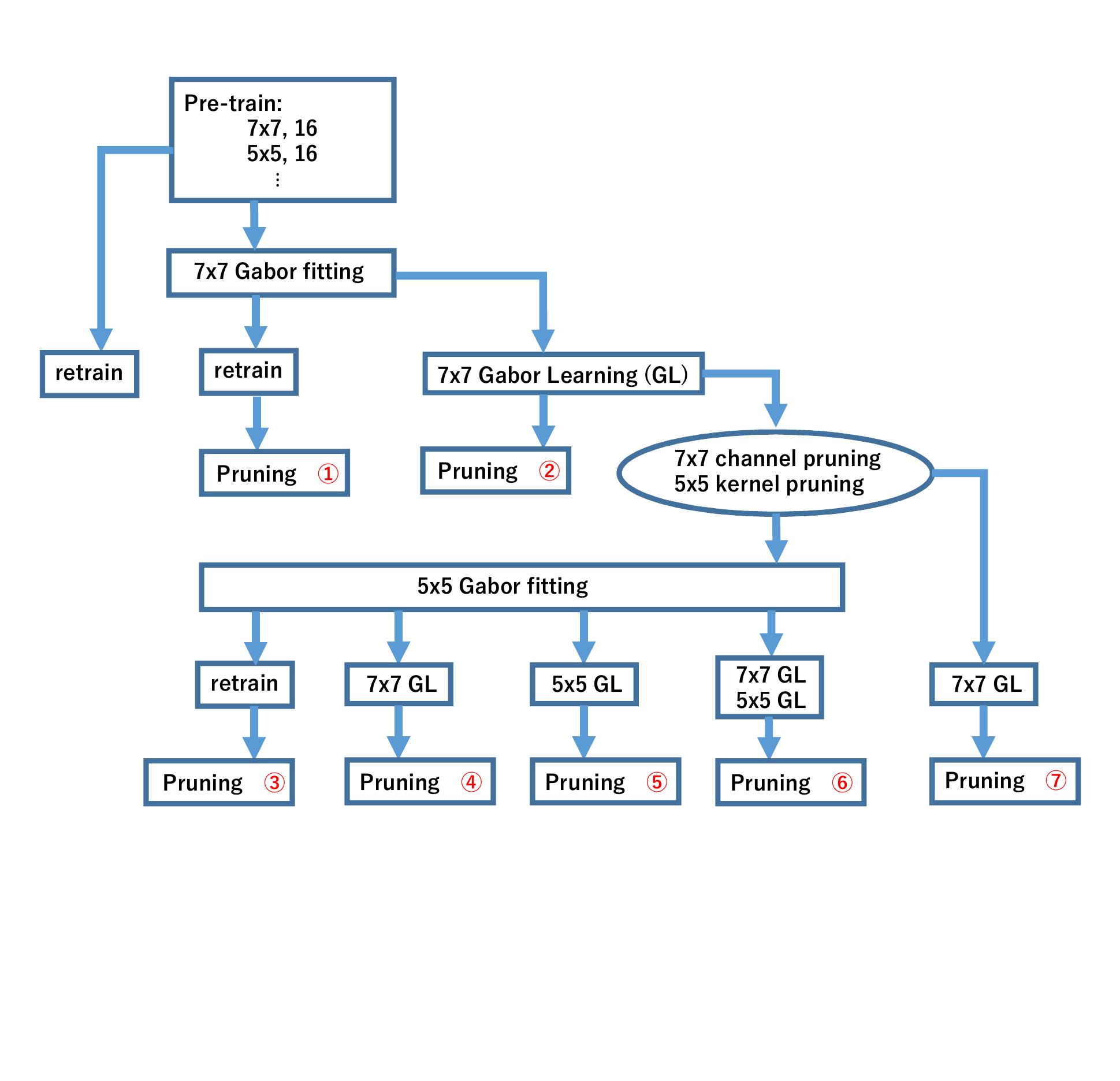}
  \vspace{-1.5cm}
  \caption{Schematic of second layer Gabor learning}
  \label{fig:scheme2}
\end{figure}

\begin{table}[t]
  \centering
  \caption{Second layer prunings}
  \label{tab:resNet7_5_accuracy}
  \renewcommand{\arraystretch}{1.3}
  \begin{tabular}{|c|c|c|c||c|}
    \hline
    \multicolumn{2}{|c|}{\,} & kernel & level & pruned\\
    \hline
    \multicolumn{5}{|c|}{First run}\\
    \hline
    \multirow{4}{*}{\makecell{7 $\times$ 7\\ fitting}} & \multirow{2}{*}{retrain \textcircled{\raisebox{-0.9pt}{1}}} & \multirow{2}{*}{\makecell{7 $\times$ 7 }} & kernel  & 10\% $\pm$ 3\%\\
    \cline{4-5}
    & & & channel & 5\% $\pm$ 5\%\\
    \cline{2-5}
    & \multirow{2}{*}{GL \textcircled{\raisebox{-0.9pt}{2}}} & \multirow{2}{*}{\makecell{7 $\times$ 7 }}  & kernel& 63\% $\pm$ 11\%\\
    \cline{4-5}
    & & & channel & 35\%$\pm$ 3\%\\
    \hline
    \hline
    \multicolumn{5}{|c|}{Second run}\\
    \hline
    \multirow{16}{*}{\makecell{5 $\times$ 5\\ fitting\\ \\ (7 $\times$ 7\\ results \\ from\\ GL)}} & \multirow{4}{*}{\makecell{7 $\times$ 7\\ retrain \\ 5 $\times$ 5\\ retrain \textcircled{\raisebox{-0.9pt}{3}}}} & \multirow{2}{*}{7 $\times$ 7 }& kernel  & 50\%$\pm$ 2\%\\
    \cline{4-5}
    & & & channel  & 49\%$\pm$ 3\% \\
    \cline{3-5}
    & & \multirow{2}{*}{5 $\times$ 5 } & kernel & 61\%$\pm$ 4\%\\
    \cline{4-5}
    & & & channel & 51\%$\pm$ 11\%\\
    \cline{2-5}
     & \multirow{4}{*}{\makecell{7 $\times$ 7\\ GL \\ 5 $\times$ 5\\ retrain \textcircled{\raisebox{-0.9pt}{4}}}} & \multirow{2}{*}{7 $\times$ 7 }& kernel  & 78\%$\pm$ 4\%\\
    \cline{4-5}
    & & & channel  & 59\%$\pm$ 3\% \\
    \cline{3-5}
    & & \multirow{2}{*}{\makecell{5 $\times$ 5 }} & kernel  & 59\%$\pm$ 2\%\\
    \cline{4-5}
    & & & channel  & 41\%$\pm$ 8\%\\
    \cline{2-5}
    & \multirow{4}{*}{\makecell{7 $\times$ 7\\ retrain \\ 5 $\times$ 5\\ GL \textcircled{\raisebox{-0.9pt}{5}}}} & \multirow{2}{*}{7 $\times$ 7 }& kernel  & 70\%$\pm$ 5\%\\
    \cline{4-5}
    & & & channel  & 66\%$\pm$ 7\%\\
    \cline{3-5}
    & & \multirow{2}{*}{\makecell{5 $\times$ 5 }} & kernel  &94\%$\pm$ 1\%\\
    \cline{4-5}
    & & & channel  &  71\%$\pm$ 13\%\\
    \cline{2-5}
    & \multirow{4}{*}{\makecell{7 $\times$ 7\\ GL \\ 5 $\times$ 5\\ GL \textcircled{\raisebox{-0.95pt}{6}}}} & \multirow{2}{*}{7 $\times$ 7 }& kernel  & 83\%$\pm$ 3\%\\
    \cline{4-5}
    & & & channel& 60\%$\pm$ 6\%\\
    \cline{3-5}
    & & \multirow{2}{*}{\makecell{5 $\times$ 5 }} & kernel  & 94\%$\pm$ 1\%\\
    \cline{4-5}
    & & & channel & 63\%$\pm$ 6\% \\
    \cline{2-5}
    \hline
    \multirow{4}{*}{\makecell{5 $\times$ 5\\ no \\fitting}}& \multirow{4}{*}{\makecell{7 $\times$ 7\\ GL \\ 5 $\times$ 5\\ retrain \textcircled{\raisebox{-0.9pt}{7}}}} & \multirow{2}{*}{7 $\times$ 7 }& kernel  & 79\%$\pm$ 5\%\\
    \cline{4-5}
    & & & channel  & 56\%$\pm$ 6\% \\
    \cline{3-5}
    & & \multirow{2}{*}{\makecell{5 $\times$ 5 }} & kernel  & 61\%$\pm$ 7\%\\

    \cline{4-5}
    & & & channel  & 41\%$\pm$ 8\%\\
    \hline 
  \end{tabular}
\end{table}

We have shown that learning the Gabor parameters in the first layer chooses fewer features.  
We expand this idea to the second layer.
The schematic of the training and pruning is in Fig. \ref{fig:scheme2}.  Each number in the figure indicates the experiments in the following table.

The first five layers of ResNet-20 are all $3 \times 3$ kernels with 16 output channels, which we denote $3 \times 3, 16$.  We expanded the kernels and removed layers to make it to two layers of $7 \times 7, 16$, and $5 \times 5, 16$ without shortcut connection.  
We trained the network four times, with the same hyperparameters as the previous run, and used this as a reference.  Note, the original ResNet-20 has an accuracy of $92.38 \pm 0.18$ after four runs, and it also shows convergence.
The Gabor fitting is the same as in previous cases, except the scale $a$, which is scaled with each kernel's maximum absolute value. 

We first prune unnecessary $7 \times 7 \times 3$ input channels from Gabor learnings. To do so, we also pruned kernels in the second layer.  Then, we repeat the Gabor fitting for $5 \times 5$ kernels.  

Table \ref{tab:7x7_5x5_2} shows the accuracy of all the combinations of learnings with 5 $\times$ 5 Gabor fitting after the first Gabor learning.  
All of them have similar results, i.e., less than 0.4\% difference.
Table \ref{tab:resNet7_5_accuracy} shows the total number of average pruning after the second run.
Their results include the pruning of $7 \times 7 \times 3$ channels in the first Gabor learnings, and it shows that further prunings of the first layer are possible.  In the second layer, input features are no longer correlated to each other like in the case of RGB inputs; hence the standard deviation of prunable channels is large.    

Using second Gabor learning in either the first or the second layer, \textcircled{\raisebox{-0.9pt}{4}}, \textcircled{\raisebox{-0.9pt}{5}}, and \textcircled{\raisebox{-0.9pt}{6}},  makes significantly more prunable channels/kernels in the first layer. It might be because the first Gabor learning makes the first layer close enough to Gabor filters. 

To check the influence on Gabor fitting in the second layer, we retrained the network without 5 $\times$ 5 Gabor fitting and compared it with the network with 5 $\times$ 5 Gabor fitting, i.e., comparing \textcircled{\raisebox{-0.9pt}{4}} and \textcircled{\raisebox{-0.9pt}{7}}. It shows that 5 $\times$ 5 Gabor fitting does not make a clear difference in the number of prunable kernels without Gabor learning.  It implies that the structure of parameter space, not just the initial values, is essential for learning.

Same as the first layer, 5 $\times$ 5 Gabor learning, \textcircled{\raisebox{-0.9pt}{5}}, and \textcircled{\raisebox{-0.95pt}{6}}, can make significantly more prunable channels/kernels.  It could be possible to extend Gabor learning to further layers.

\section{CONCLUSION AND DISCUSSION}
\label{sec:prior}
In this paper, we propose to retrain networks by Gabor learning in the earlier layers of CNNs to decrease the number of features/channels.  Using strong inductive biases by restricting features to Gabor filters help networks to choose specific filters.  As with classical image processing, we showed choosing the right features for filters can make a smaller number of filters after learning.  

Our method does not compete with the other compressing methods since it does not alter the structure of the networks.  Hence, it can be used as a good preprocessing for network compression.


\bibliographystyle{IEEEtran}
\bibliography{IEEEabrv,mybibfile}

\end{document}